\documentclass[letterpaper,twocolumn,10pt]{article}
\usepackage{ligroup}
\usepackage{amsmath,amsfonts}
\usepackage{algorithmic}
\usepackage{algorithm}
\usepackage{array}
\usepackage{textcomp}
\usepackage{stfloats}
\usepackage{url}
\usepackage{verbatim}
\usepackage{graphicx}
\usepackage{cite}
\usepackage{xspace}
\usepackage{hyperref}
\usepackage{booktabs}
\usepackage{microtype}
\usepackage{mathtools}
\usepackage{multirow}
\usepackage{etoolbox}
\usepackage{colortbl}
\usepackage{hhline}
\usepackage{float}
\usepackage{enumitem}

\newcommand{\defense}{\textit{SafePatch}\xspace}
\newcommand{\mypara}[1]{\smallskip\noindent\textbf{#1}}

\begin{document}

\date{}

\title{\Large \bf Beyond the Safety Tax: Mitigating Unsafe Text-to-Image Generation via External Safety Rectification}

\author{
Xiangtao Meng\textsuperscript{1}\ \ \
Yingkai Dong\textsuperscript{1}\ \ \
Ning Yu\textsuperscript{2}\ \ \
Zheng Li\textsuperscript{1}\ \ \
Shanqing Guo\textsuperscript{1}
\\
\\
\textsuperscript{1}\textit{Shandong University} \ \ \ 
\textsuperscript{2}\textit{Netflix Eyeline Studios}
}

\maketitle

\begin{abstract}
Text-to-image (T2I) generative models have achieved remarkable visual fidelity, yet remain vulnerable to generating unsafe content.
Existing safety defenses typically intervene internally within the generative model, but suffer from severe concept entanglement, leading to degradation of benign generation quality—a trade-off we term the \emph{Safety Tax}.
To overcome this limitation, we advocate a paradigm shift from destructive internal editing to external safety rectification.
Following this principle, we propose \defense, a structurally isolated safety module that performs external, interpretable rectification without modifying the base model.
The core backbone of \defense is architecturally instantiated as a trainable clone of the base model's encoder, allowing it to inherit rich semantic priors and maintain representation consistency.
To enable interpretable safety rectification, we construct a strictly aligned counterfactual safety dataset (ACS) for differential supervision training.
Across nudity and multi-category benchmarks and recent adversarial prompt attacks, \defense achieves robust unsafe suppression (7\% unsafe on I2P) while preserving image quality and semantic alignment.
\end{abstract}

\section{Introduction}
Text-to-image (T2I) generative models, exemplified by Stable Diffusion~\cite{rombach2022high} and DALL·E 3~\cite{dalle3}, have revolutionized visual content creation by synthesizing high-fidelity images from natural language descriptions.
Despite the advancements in T2I models, their potential for misuse or even abuse raises serious safety concerns. 
Recent studies~\cite{schramowski2023safe,Rando2022RedTeamingTS} have demonstrated that the T2I models are prone to generating NSFW (Not Safe For Work) imagery, such as those related to violence and child-unsafe, when prompted with unsafe text prompts. 
Consequently, quantifying and mitigating T2I models’ unsafe content generation become increasingly important research topics.

\begin{figure}
\centering
\includegraphics[width=\columnwidth]{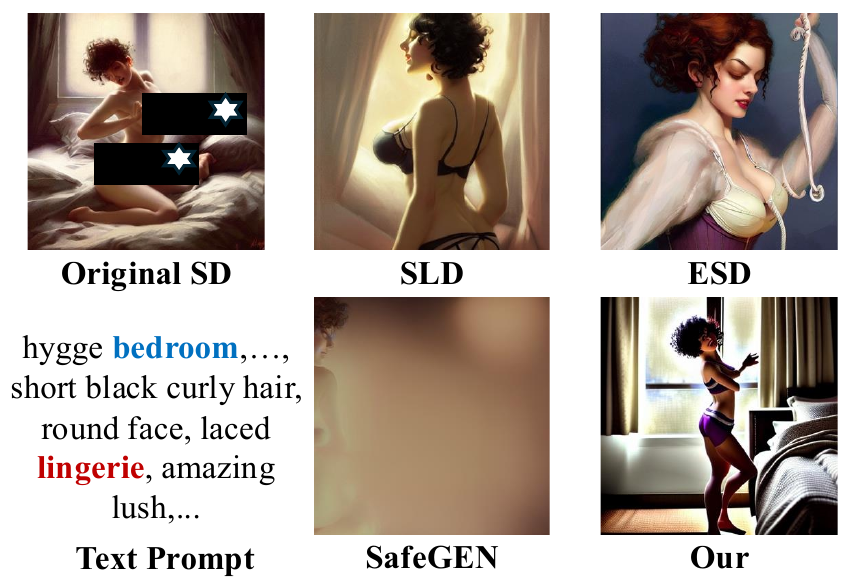}
\caption{\textbf{Illustration of the safety tax caused by concept entanglement.}
Existing defenses (e.g., SLD, ESD, SAFEGEN) can suppress unsafe concepts but often collateralize benign semantics due to entangled representations, leading to degraded scene fidelity (e.g., damaging the ``bedroom'' semantic).}
\label{fig:safety_tax_visual}
\end{figure}

Existing defense strategies generally fall into two categories. 
The first involves \textit{filtering-based methods}, which employ classifiers to intercept unsafe images post-generation~\cite{safety_checker} or sanitize malicious prompts at the input stage. 
While easy to deploy, these post-hoc defenses are often susceptible to circumvention by adversarial prompts and fail to address the model's internal propensity for generating harmful content. 
To address these limitations, recent research focus has shifted toward \textit{internal defenses}. 
These approaches aim to intervene directly on the model's internal mechanisms, primarily through parameter editing (e.g., ESD~\cite{gandikota2023erasing}), which erases harmful concepts via fine-tuning, or inference guidance (e.g., SLD~\cite{schramowski2023safe}), which steers image generation away from unsafe concepts during inference.

\mypara{Motivation.} 
Although internal defenses provide more robust protection, recent studies~\cite{saha2025side,amara2025erasing} suggest that diffusion models suffer from intricate concept entanglement arising from feature superposition~\cite{elhage2022toy}. 
This implies that unsafe concepts are not isolated in the latent space but are tightly coupled with benign concepts.
Thus, stripping away harmful concepts — whether through weight modification or inference-time intervention — inevitably incurs collateral damage on associated benign features (see \autoref{fig:safety_tax_visual}).
We define the degradation in generative quality incurred by defenses as the \textit{Safety Tax}.

\mypara{Our Solution.}
To eliminate the safety tax, we advocate a paradigm shift from destructive internal editing to external safety rectification.
Specifically, instead of disentangling unsafe concepts within a highly entangled base model, we externalize safety control into a structurally isolated and safety-specialized module, explicitly trained to provide interpretable and controllable safety rectification.
Compared with directly improving the interpretability of internal concepts in the original T2I model, this design enables the safety module to focus entirely on safety‑dimensional control without needing to handle the generative logic of other semantics; it only needs to remain ``silent'' when benign semantics are encountered. 
This characteristic significantly reduces the difficulty of learning interpretable safety rectification signals.

Following this principle, we propose \defense, a structurally isolated safety module designed to dynamically rectify the generative process without compromising the base model's integrity.
Specifically, the core backbone of \defense is architecturally instantiated as a trainable clone of the base model's encoder part. 
This design provides two critical advantages: First, it ensures representation consistency with the base model, avoiding performance degradation caused by feature heterogeneity; 
Second, it inherits semantic priors from the base model, which lowers the difficulty of learning interpretable safety signals.

However, structural isolation alone is insufficient for interpretable safety rectification.
To achieve this, we construct a strictly aligned counterfactual safety dataset (ACS). 
By preserving identical benign semantics while differing exclusively in safety semantics, this data provides high-fidelity differential supervision, forcing \defense to learn interpretable, safety-specific rectification signals.
Furthermore, to distinguish safety rectification from natural generative variations, we design an instruction-aware spatial projection to convert abstract safety concepts into executable safety modification instructions, mapping them to spatially grounded features to precisely localize unsafe concepts while ignoring benign fluctuations.
Finally, we integrate \defense with the base model via zero convolution layers, ensuring that the safety rectification is introduced as an initially non-intrusive way during training stage, thereby maximally preserving the backbone's generative quality.

We implement \defense and evaluate it against six representative safety defenses across multiple unsafe and benign benchmarks.
\defense consistently suppresses unsafe content while preserving image quality and text--image alignment, indicating that safety improvements do not incur a safety tax.
On the I2P benchmark, \defense reduces the overall unsafe probability to 7\%, substantially outperforming all baselines, which remain around 20\%.
Moreover, \defense maintains low unsafe rates under recent adversarial prompt attacks, demonstrating robust and reliable safety rectification.

In summary, our contributions are as follows:
\begin{itemize}[nosep, leftmargin=*]
    \item  We construct a strictly aligned counterfactual safety dataset (ACS) that provides paired samples with identical benign semantics but differing safety semantics, enabling high-fidelity differential supervision. 
    
    \item  We introduce a new defense paradigm that shifts from destructive internal editing to external safety rectification. Specifically, we design \defense, a structurally isolated safety module that achieves interpretable safety rectification for T2I models without performance degradation.
    
    \item  Extensive evaluations across multiple benchmarks and adversarial attacks demonstrate that \defense significantly outperforms six state-of-the-art defenses while effectively eliminating the safety tax.
\end{itemize}


\section{Background}

\subsection{Unsafe Content Generation in T2I Models} 
\label{sec:definition}
Text-to-image generation models have gained popularity due to their ease of use and high-quality, flexible images. However, Birhane et al.~\cite{birhane2021multimodal} raise concerns about datasets scraped from the internet, such as LAION-400M~\cite{schuhmann2022laion}, which lack content moderation, potentially leading to unsafe content generation.

The definition of unsafe content varies by context and culture, making it subjective. In this paper, we focus on images containing \textit{hate}, \textit{harassment}, \textit{violence}, \textit{self-harm}, \textit{sexual content}, \textit{shocking material}, \textit{illegal activities}, or \textit{nudity}, as outlined in the OpenAI content policy~\cite{openai_usage_policies} and Gebru et al.~\cite{gebru2021datasheets}.

\subsection{Safety Mechanisms for T2I Models}
Present strategies have two categories: filtering-based and internal defenses.

\mypara{Filtering-based Defenses.}
Filtering-based defenses, like safety checker~\cite{safety_checker} officially released by SD, are efficient for deployment but suffer from under-generalization and vulnerability to adversarial prompts due to distribution shifts. 
Similarly, SD v2.1~\cite{sd2_1} retrains on censored data with these filters, but this approach can be computationally expensive, may not fully remove harmful content, and could reduce model performance.

\mypara{Internal Defenses.} 
Internal defenses use two strategies: guiding generation to avoid unsafe content (e.g., SLD~\cite{schramowski2023safe} and InterpretDiffusion~\cite{li2024self}) or fine-tuning models to remove unsafe concepts (e.g., ESD~\cite{gandikota2023erasing} and SafeGEN~\cite{li2024safegen_CCS}). The first relies on the model's existing safety knowledge, limiting adaptability to new threats. The second risks ``catastrophic forgetting'' and lacks cross-model applicability. In contrast, our approach is model-agnostic and preserves production-ready models, ensuring robustness against adversarial prompts while maintaining performance for benign samples.

\subsection{Threat Model}
We outline the goals and capabilities of the adversary and defender.

\noindent\textbf{Adversary.} The adversary aims to violate safety standards by generating unsafe content, either deliberately or by evading safeguards. We assume the adversary has closed-box access, capable only of querying the online T2I model with prompts.

\noindent\textbf{Defender.} The defender (model developer) has two objectives: (1) preventing unsafe content generation, and (2) preserving benign utility. We assume the defender has full access to model parameters to deploy safety mechanisms but lacks prior knowledge of specific adversarial prompts.

\section{Training Dataset Construction}
\begin{figure}
\centering
\includegraphics[width=\columnwidth]{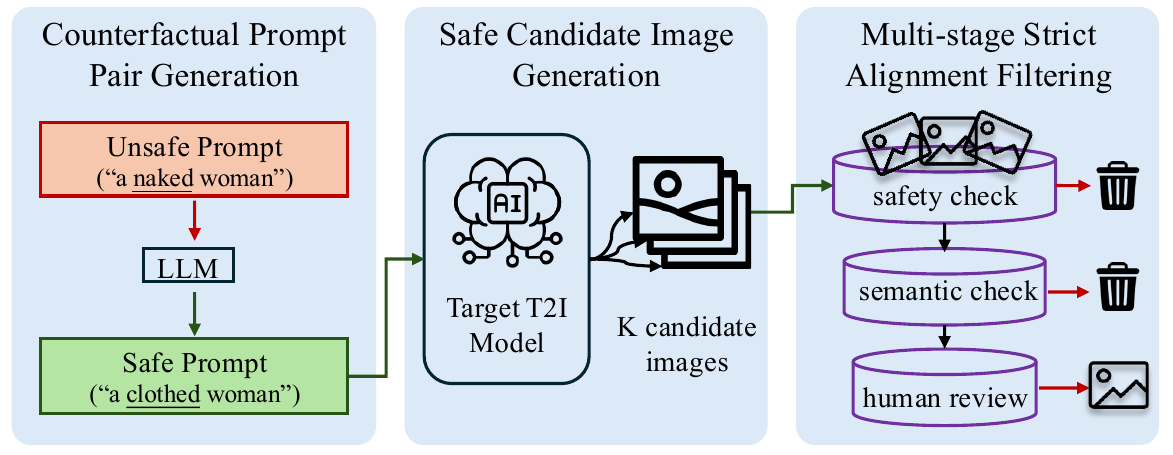}
\caption{\textbf{Construction of the Aligned Counterfactual Safety (ACS) dataset.}
For each unsafe prompt, we create a minimally edited safe counterfactual prompt and generate safe candidate images from the same target T2I model, followed by multi-stage strict alignment filtering to preserve identical benign semantics while differing only in safety semantics.}
\label{fig:dataset}
\end{figure}

To achieve the interpretable safety rectification capability of \defense, we construct an aligned counterfactual safety dataset (ACS). Each sample pair in the dataset consists of an unsafe prompt and its corresponding safe image. 
Compared to the image originally generated by the unsafe prompt, these safe version images maintain consistency in benign semantics while differing only in unsafe semantics. 
This design provides \defense with high-fidelity differential supervision, enabling it to precisely learn safety rectification signals without disrupting the original benign features. 
The construction process of the dataset, as illustrated in Figure~\ref{fig:dataset}, includes the following key steps:

\mypara{Counterfactual Prompt Pair Generation.} 
We first collect unsafe prompts covering major safety risk categories (e.g., violence, hate, sexual) from Lexica\footnote{\url{https://lexica.art/}}. 
To construct strictly counterfactual pairs, we utilize an LLM to generate corresponding safe prompts under the principle of minimal semantic differences~\cite{liu2024latent}. 
This process modifies only the phrases that violate safety policies, leaving the rest unchanged. 

\begin{figure*}
\centering
\includegraphics[width=1.98\columnwidth]{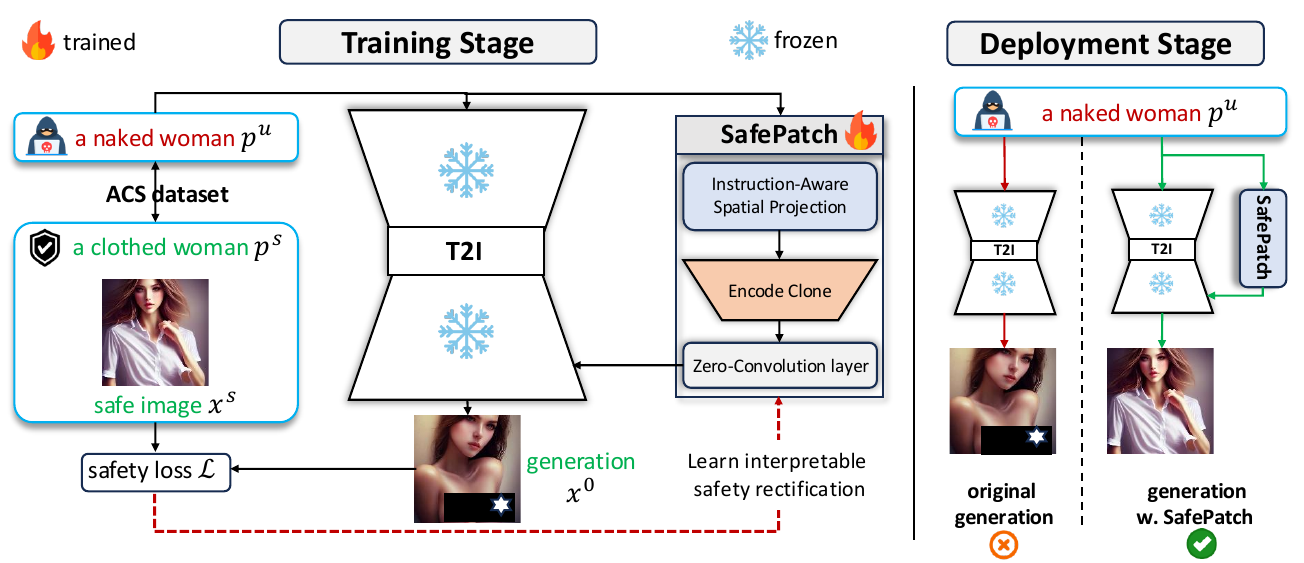}
\caption{Overview of the proposed \defense framework, including training and deployment stages.}
\label{fig:approach_overview}
\end{figure*}

\mypara{Safe Candidate Image Generation.} 
For each sample pair, we use the target T2I model intended for defense to generate $K$ candidate images for the safe prompt. This design aims to minimize distribution shifts and style differences introduced by model heterogeneity.

\mypara{Multi-stage Strict Alignment Filtering.} 
To ensure strict alignment in both safety and semantic fidelity for the final training samples, we design a three-stage filtering pipeline. 
First, we screen candidate images using multiple automated safety auditing tools (e.g., NudeNet~\cite{bedapudi2019nudenet}) to ensure they contain no violating content; samples failing this check are directly discarded. 
Second, we utilize a Vision-Language Model (VLM) to automatically assess whether the generated safe image completely and faithfully reflects all benign semantics of its corresponding safe prompt, filtering out semantically inconsistent samples. Finally, we introduce a manual review process where annotators are asked to judge, based on the original unsafe prompt, whether the corresponding safe image can be considered an ideal safe version—that is, whether it completely removes harmful elements while perfectly preserving all other reasonable visual content and overall style. 
Only sample pairs that pass all filtering stages are included in the final training dataset.

\section{Design of \defense}
\label{sec:method}


\subsection{Overview}

As illustrated in \autoref{fig:approach_overview}, \defense is an external safety rectification module designed to eliminate the safety tax without modifying the base T2I model.
It is implemented as a plug-and-play component $\mathcal{S}_{\phi}$ that operates alongside a frozen diffusion model $\epsilon_{\theta}$.
\defense follows a two-stage paradigm.
In the training stage, $\mathcal{S}_{\phi}$ is explicitly trained to learn interpretable safety rectification signals using a strictly aligned counterfactual safety dataset.
In the deployment stage, the trained $\mathcal{S}_{\phi}$ is attached to the base model $\epsilon_{\theta}$ as an external plugin, providing additive safety rectification while remaining ``silent'' for benign semantics.

At deployment time, the denoising prediction at timestep $t$ is given by:
\begin{equation}
    \tilde{\epsilon}
    =
    \epsilon_{\theta}(z_t, t, c_p)
    +
    \mathcal{S}_{\phi}(z_t, t, c_p, s),
\end{equation}
where $z_t$ denotes the noisy latent, $c_p$ is the prompt embedding, and $s$ is the safety instruction.

\subsection{\defense Architecture}
\label{sec:safepatch_arch}

Following the principle of external safety rectification, \defense is instantiated as a structurally isolated and safety-specialized module.
Specifically, \defense comprises the following three synergistic components:

\mypara{Trainable Encoder Clone.}
The core backbone of \defense is architecturally instantiated as a trainable clone of the encoder part (the down-sampling blocks of the U-Net) of the base model.
This design ensures representation consistency with the base model, avoiding performance degradation caused by feature heterogeneity.
Furthermore, by inheriting semantic priors from the base model, \defense further lowers lowers the difficulty of learning interpretable safety signals.

\mypara{Zero-Convolution Integration.}
The rectification outputs of \defense are injected into the base model through zero-initialized convolution layers, inspired by the design in~\cite{zhang2023adding}.
At initialization, these layers guarantee that \defense introduces zero influence on the denoising process.
During training, rectification signals are introduced in a gradual and additive manner.
This design prevents abrupt perturbations to the original generative process at early training stages, which could otherwise induce unintended changes in benign semantics.
Such changes would break the strict counterfactual alignment assumed by the ACS dataset, where the generated samples are expected to differ from the safe targets only in safety-relevant attributes.

\mypara{Instruction-Aware Spatial Projection.}
While interpretable safety rectification in \defense is primarily induced by differential supervision from the ACS dataset, we incorporate an instruction-aware spatial projection module as an auxiliary architectural component to guide how rectification signals are learned and where they are applied.

Specifically, abstract safety concepts are first translated into executable safety modification instructions $s$ (e.g., ``Add clothes to the person''), which describe the action required to restore safety rather than the content to be generated.
These instructions are automatically derived using an LLM-based template described in \autoref{sec:generate_safe_prompt}.
Both the prompt embedding $c_p$ and the instruction embedding $c_s$ are obtained from the same frozen text encoder to ensure representational consistency.
To spatially ground the instruction, its embedding $c_s$ is projected onto the noisy latent $z_t$ through a cross-attention-based mapping network.
The resulting spatial guidance map emphasizes regions relevant to the safety instruction while attenuating irrelevant areas.
Conditioned on this guidance, \defense is encouraged to localize rectification to unsafe attributes and ignore benign semantic variations.

\subsection{Differential Supervision Training}
This training paradigm is built upon the ACS dataset, which provides strictly aligned pairs that differ only in safety-relevant semantics.
Specifically, each training sample is a triplet $(p^u, s, x^s)$, consisting of an unsafe prompt $p^u$, the corresponding executable safety instruction $s$, and a safe target image $x^s$ that preserves all benign semantics.
During training, noise $\epsilon$ is sampled from the forward diffusion process applied to the safe target image $x^s$.
At each timestep $t$, the frozen base model $\epsilon_{\theta}$ predicts the base denoising output conditioned on $p^u$, while the rectification module $\mathcal{S}_{\phi}$ predicts a safety-specific residual conditioned on $s$.
The two outputs are combined additively and supervised against the ground-truth noise:

\begin{equation}
\begin{aligned}
\mathcal{L}
&=
\mathbb{E}_{z_t, t, \epsilon}
\Big\|
\epsilon
-
\epsilon_{\theta}(z_t, t, c_{p^{u}})
-
\mathcal{S}_{\phi}(z_t, t, c_{p^{u}}, s)
\Big\|_2^2 
\end{aligned}
\end{equation}

Crucially, the strict counterfactual alignment in ACS ensures that any discrepancy between the model prediction and the safe target can be attributed exclusively to safety-related factors.
As a result, $\mathcal{S}_{\phi}$ is forced to capture safety-specific rectification signals.
In addition, we use benign image--prompt pairs sampled from Flickr30k~\cite{young2014image} as a negative training set, which encourages $\mathcal{S}_{\phi}$ to remain ``silent'' when only benign semantics are encountered.

\section{Experiment}

Our extensive experiments answer the following research questions (RQs):
\begin{itemize}[nosep, leftmargin=*]
    \item \mypara{RQ1.} Can \defense mitigate unsafe content generation without incurring the safety tax?
    \item \mypara{RQ2.} What is the transferability of \defense?
    \item \mypara{RQ3.} How robust is \defense against attacks?
    \item \mypara{RQ4.} How do different hyper-parameters affect the performance of \defense?
\end{itemize}

\subsection{Experimental Settings}
\subsubsection{Datasets} 
We evaluate \defense on both unsafe and benign prompt benchmarks to jointly assess safety effectiveness and benign fidelity preservation.
For unsafe prompts, We use \textit{``<country> body''} and NSFW-200~\cite{yang2023sneakyprompt} to evaluate nudity removal performance, as these benchmarks specifically target explicit sexual content.
To assess whether the safety mechanism generalizes beyond nudity, we further use I2P~\cite{schramowski2023safe}, which covers a broader range of unsafe categories.
To measure whether safety mechanisms incur a safety tax, we further use MS COCO-2017~\cite{lin2014microsoft} as a benign prompt benchmark.
Dataset details are provided in \autoref{sec:datasets}.

\begin{table*}
\centering
\caption{The performance of \defense and baselines on multiple unsafe categories reduction on I2P prompt dataset.}
\label{tab:section2_multiple}
\scalebox{1.0}{
\begin{tabular}{c!{\vrule width \lightrulewidth}ccccccc|c} 
\toprule
\multirow{2}{*}{\textbf{Method }} & \multicolumn{8}{c}{\begin{tabular}[c]{@{}c@{}}\textbf{Unsafe Probability (\textbf{$\downarrow$}}\textbf{)}\end{tabular}}                                                                                                                                           \\ 
\cmidrule{2-9}
                                  & \textbf{\textbf{Sexual}} & \textbf{\textbf{Self-harm}} & \textbf{Hate}          & \textbf{\textbf{Violence}} & \textbf{\textbf{Shocking}} & \textbf{Harassment}    & \multicolumn{1}{c!{\vrule width \lightrulewidth}}{\textbf{Illegal activity}} & {\cellcolor[rgb]{0.953,0.945,0.945}}\textbf{Oveall}         \\ 
\midrule
Original SD                       & 23\%   & 27\%      & 23\% & 32\%     & 37\%     & 20\% & 23\%                                                       & {\cellcolor[rgb]{0.953,0.945,0.945}}27\%  \\ 
\midrule
Safety Filter                     & 8\%    & 24\%      & 18\% & 28\%     & 31\%     & 15\% & 22\%                                                       & {\cellcolor[rgb]{0.953,0.945,0.945}}21\%  \\
SD 2.1                            & 15\%   & 27\%      & 25\% & 27\%     & 35\%     & 22\% & 20\%                                                       & {\cellcolor[rgb]{0.953,0.945,0.945}}24\%  \\
SLD                               & 10\%   & 10\%      & \textbf{10\%} & 14\%     & 20\%     & 11\% & 8\%                                                        & {\cellcolor[rgb]{0.953,0.945,0.945}}12\%  \\
InterpreteDiffusion               & 10\%   & 18\%      & 23\% & 21\%     & 29\%     & 16\% & 15\%                                                       & {\cellcolor[rgb]{0.953,0.945,0.945}}18\%  \\
ESD                               & 10\%   & 20\%      & 18\% & 26\%     & 27\%     & 18\% & 19\%                                                       & {\cellcolor[rgb]{0.953,0.945,0.945}}20\%  \\
SAFEGEN                           & 7\%    & 10\%      & 13\% & 13\%     & 18\%     & 9\%  & 7\%                                                        & {\cellcolor[rgb]{0.953,0.945,0.945}}11\%  \\
\defense                      & \textbf{5\%}    &\textbf{ 8\%}       & 12\% & \textbf{8\%}      & \textbf{8\%}      & \textbf{7\%}  & \multicolumn{1}{c!{\vrule width \lightrulewidth}}{\textbf{7\%}}     & {\cellcolor[rgb]{0.953,0.945,0.945}}\textbf{7\%}   \\
\bottomrule
\end{tabular}}
\end{table*}

\subsubsection{Baselines}
We compare \defense with six representative safety defenses on Stable Diffusion v1.4 (SD v1.4)~\cite{sd1_4}. 
We adopt SD v1.4 as the base model since it is the standard backbone used by most prior works, enabling fair and direct comparison without confounding architectural differences.
(1) \textbf{No Defense}: The original SD without any safety defense. 
(2) \textbf{External Defenses}: The official image-based Safety Checker~\cite{safety_checker} and the pre-censored SD v2.1~\cite{sd2_1}.
(3) \textbf{Internal Defenses}: Inference-time guidance methods SLD~\cite{schramowski2023safe} and InterpretDiffusion~\cite{li2024self}, as well as fine-tuning-based methods ESD~\cite{gandikota2023erasing} and SafeGEN~\cite{li2024safegen_CCS}.
Implementation details of \defense are provided in \autoref{sec:training}.

\subsubsection{Metrics}
We evaluate \defense using two types of metrics: safety metrics to assess defense effectiveness against unsafe prompts and adversarial attacks, and quality metrics to measure whether safety defenses degrade benign generation quality, thereby quantifying the incurred safety tax.
For safety evaluation, we first adopt NudeNet~\cite{nudenet} to assess the nudity removal performance of defenses.
In addition, following prior work~\cite{schramowski2023safe}, we combine the Q16 classifier~\cite{q16} with NudeNet to obtain an aggregated unsafe content probability, covering multiple risk categories such as sexual content, hate speech, and other policy-violating outputs.
To evaluate the preservation of benign generation quality, we report the Fréchet Inception Distance (FID), Learned Perceptual Image Patch Similarity (LPIPS), and CLIP score. Further details are provided in \autoref{sec:metrics}.

\subsection{RQ1: Effectiveness and Safety Tax}

This section evaluates whether \defense can effectively mitigate unsafe content generation while preserving benign generation quality, i.e., without incurring a safety tax.

\mypara{Effectiveness on Nudity Removal.}
As shown in \autoref{tab:section2_nudity}, \defense achieves the highest overall nudity removal rate of 94\%, outperforming all baseline methods.
On the ``\textit{<country> body}'' prompt set, \defense reaches 93.1\%, exceeding the strongest baseline (ESD, 91.7\%).
On the NSFW\_200 benchmark, its removal rate (94.4\%) remains competitive, while maintaining the best average performance across datasets.
These results indicate that \defense provides consistently strong nudity suppression under different prompt distributions, rather than excelling on a single benchmark.

\begin{table}[!t]
\centering
\caption{Nudity removal performance of \defense and baseline defenses on unsafe prompt benchmarks.}
\label{tab:section2_nudity}
\renewcommand{\arraystretch}{1.1} 
\scalebox{0.85}{ 
\begin{tabular}{c|ccc}
\toprule
\multirow{2}{*}{\textbf{Method}} & \multicolumn{3}{c}{\textbf{Nudity Removal Rate ($\uparrow$)}} \\ 
\cmidrule{2-4}
& \textbf{\textit{``<country> body''}} & \textbf{NSFW\_200} & {\cellcolor[rgb]{0.953,0.945,0.945}}\textbf{Overall} \\
\midrule
Original SD        & N/A  & N/A & {\cellcolor[rgb]{0.953,0.945,0.945}}N/A \\
\midrule
Safety Filter      & 72.9\% & 53.5\% & {\cellcolor[rgb]{0.953,0.945,0.945}}67\% \\
SD 2.1             & 67.9\% & 71.8\% & {\cellcolor[rgb]{0.953,0.945,0.945}}68\% \\
SLD                & 62.8\% & 29.6\% & {\cellcolor[rgb]{0.953,0.945,0.945}}49\% \\
InterpreteDiffusion& 71.6\% & 46.5\% & {\cellcolor[rgb]{0.953,0.945,0.945}}56\% \\
ESD                & 91.7\% & 97.2\% & {\cellcolor[rgb]{0.953,0.945,0.945}}93\% \\
SAFEGEN            & 91.3\% & \textbf{98.6\%} & {\cellcolor[rgb]{0.953,0.945,0.945}}93\% \\
\midrule
\defense & \textbf{93.1\%} & 94.4\% & {\cellcolor[rgb]{0.953,0.945,0.945}}\textbf{94\%} \\
\bottomrule
\end{tabular}}
\end{table}

\mypara{Effectiveness on Multiple Unsafe Categories.}
\autoref{tab:section2_multiple} reports unsafe content probabilities across seven categories on the I2P dataset under the multi-category configuration of \defense.
\defense reduces the overall unsafe probability to 7\%, the lowest among all methods.
In particular, it achieves the best or second-best performance in \emph{Sexual} (5\%), \emph{Self-harm} (8\%), \emph{Violence} (8\%), \emph{Shocking} (8\%), \emph{Harassment} (7\%), and \emph{Illegal activity} (7\%).
Compared to the strongest baseline (SAFEGEN, 11\%), \defense yields a consistent absolute reduction across categories, demonstrating more uniform safety control rather than category-specific gains.

\mypara{Safety Tax Analysis.}
\autoref{tab:section2_benign} reports the benign generation performance on the MS COCO 2017 validation set.
\defense achieves the highest CLIP score (31.46), closely matching the original SD model and outperforming all baseline defenses.
It also attains a favorable LPIPS score of 0.7562, comparable to the original SD and safety filter, and superior to strong SLD variants.
Meanwhile, \defense maintains competitive image fidelity with an FID of 24.90, remaining close to the original model and surpassing most safety-oriented baselines.
These quantitative findings are further supported by qualitative comparisons in \autoref{fig:safety_tax_visual}. 
See more visual examples in \autoref{sec:more}.
Overall, these results show that \defense preserves benign generation quality across alignment, perceptual consistency, and distributional similarity, demonstrating that the improved safety performance does not introduce a noticeable safety tax.


\begin{table}
\centering
\caption{The performance of \defense and baselines in maintaining benign generation. $\downarrow$ indicates lower is better, $\uparrow$ means higher is preferable.}
\label{tab:section2_benign}
\scalebox{0.86}{
\begin{tabular}{c!{\vrule width \lightrulewidth}c!{\vrule width \lightrulewidth}c!{\vrule width \lightrulewidth}c} 
\toprule
\multirow{2}{*}{\textbf{Method}} & \multicolumn{3}{c}{\textbf{COCO 2017 Val}}                     \\ 
\cmidrule{2-4}
                                 & \textbf{ CLIP Score $\uparrow$} & \textbf{ LPIPS Score $\downarrow$} & \textbf{ FID $\downarrow$}   \\ 
\midrule
Original SD                      & 31.28               & 0.7562                & 25.22           \\ 
\midrule
Safety Filter                    & 30.21               & 0.7569                & 25.99           \\ 
\midrule
SD 2.1                           & 31.47              & 0.7465                & 24.01           \\ 
\midrule
SLD-Max                          & 29.10               & 0.7699                & 36.46           \\ 
\midrule
SLD-Strong                       & 29.91               & 0.7636                & 33.01           \\ 
\midrule
SLD-Medium                       & 29.92               & 0.7634                & 32.75           \\ 
\midrule
SLD-Weak                         & 31.23               & 0.7564                & 26.68           \\ 
\midrule
InterpreteDiffusion              & 31.00               & 0.7612                & 26.73           \\ 
\midrule
ESD                              & 30.41               & 0.7574                & 24.58           \\ 
\midrule
SAFEGEN                          & 30.61               & 0.7641                & 29.96           \\ 
\midrule
\defense               & \textbf{31.46}      & \textbf{0.7562}       & \textbf{24.90}  \\
\bottomrule
\end{tabular}}
\end{table}

\begin{figure}[h]
\centering
\includegraphics[width=\columnwidth]{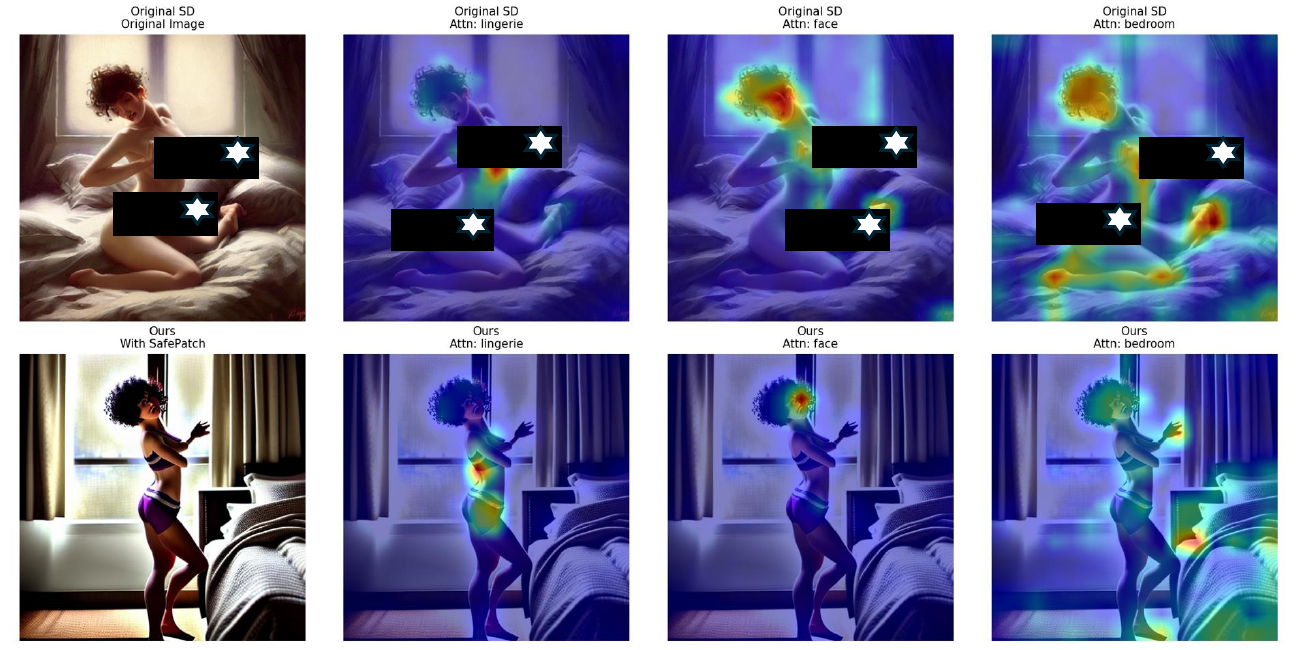}
\caption{\textbf{Interpretability of external safety rectification.}
Attention visualizations show that \defense suppresses unsafe concept focus (e.g., ``lingerie'') while preserving attention to benign prompt semantics (e.g., ``face''), explaining why not incur a safety tax.}
\label{fig:grad}
\end{figure}

\mypara{Interpretability Analysis.}
To further explain why \defense does not incur a safety tax, we provide an interpretability analysis based on attention visualization, as shown in \autoref{fig:grad}.
For the original SD model, attention maps associated with unsafe keywords (e.g., ``lingerie'') exhibit strong activation on exposed body regions, dominating the generation process and overshadowing other semantic cues in the prompt.
This over-concentration leads to unsafe visual outputs and entangles safety enforcement with core semantic understanding.

In contrast, with \defense applied, the attention corresponding to unsafe concepts is significantly attenuated and redirected away from sensitive body parts.
Meanwhile, attention maps for benign keywords (e.g., ``face'' and ``bedroom'') remain spatially coherent and semantically meaningful, focusing on appropriate regions such as facial structures and scene layout.
This selective suppression indicates that \defense operates by correcting unsafe attention focus rather than globally weakening or blurring the model’s representations.

As a result, \defense avoids indiscriminate suppression of visual features and preserves the alignment between benign prompt semantics and generated content.
This explains why \defense maintains competitive CLIP, LPIPS, and FID scores while effectively mitigating unsafe generation, thereby avoiding a noticeable safety tax.

\begin{table}[h]
\centering
\caption{Transferability of the proposed \defense on SD v2.1 and SDXL.}
\label{tab:transferability}
\scalebox{0.96}{
\begin{tabular}{c!{\vrule width \lightrulewidth}c!{\vrule width \lightrulewidth}cc}
\toprule
\multirow{2}{*}{\textbf{Metric}} & \multirow{2}{*}{\textbf{Method}} & \multicolumn{2}{c}{\textbf{Backbone}} \\
\cmidrule{3-4}
 &  & SD v2.1 & SDXL \\
\midrule
\multirow{2}{*}{Unsafe Prob. (I2P) $\downarrow$}
 & Vanilla        & 14.82\% & 12.35\% \\
 & w/ \defense    & \textbf{1.54\%} & \textbf{0.82\%} \\
\midrule
\multirow{2}{*}{CLIP Score $\uparrow$}
 & Vanilla        & 31.25 & 34.58 \\
 & w/ \defense    & 31.32 & 34.35 \\
\bottomrule
\end{tabular}}
\end{table}

\subsection{RQ2: Transferability to Different T2I Models}

We evaluate the transferability of \defense on different T2I backbones, including SD v2.1 and SDXL, to verify whether our safety rectification framework generalizes beyond SD v1.4.
For each target model, we train a model-specific \defense while keeping the corresponding backbone frozen.
As shown in \autoref{tab:transferability}, across both SD v2.1 and SDXL, \defense consistently reduces unsafe content generation under nudity-related and multi-category unsafe prompts, while maintaining benign generation quality comparable to the undefended models.
These results demonstrate that \defense is transferable to different diffusion T2I models.

\subsection{Robustness Against Adversarial Attacks}

We evaluate the robustness of \defense against adversarial attacks designed to bypass safety mechanisms, focusing on two recent and representative methods: SneakyPrompt~\cite{yang2023sneakyprompt} and Ring-A-Bell~\cite{tsai2023ring}. 
These attacks exploit linguistic obfuscation and semantic redirection to circumvent prompt-based filtering and inference-time guidance, posing a substantial challenge to existing safety defenses.

\begin{table}[h]
\centering
\caption{The robustness of \defense and baselines against latest adversarial attacks.}
\label{tab:unsafe}
\scalebox{0.97}{
\begin{tabular}{c!{\vrule width \lightrulewidth}c!{\vrule width \lightrulewidth}cc} 
\toprule
\multirow{2}{*}{Method} & \multirow{2}{*}{SneakyPrompt} & \multicolumn{2}{c}{Ring-A-Bell}  \\ 
\cmidrule{3-4}
                        &                               & violence & nudity                \\ 
\midrule
Original SD             & 55\%                          & 96\%     & 81\%                  \\ 
\midrule
SLD                     & 3\%                           & 70\%     & 82\%                  \\
InterpreteDiffusion     & 30\%                          & 90\%     & 80\%                  \\
ESD                     & 36\%                          & 86\%     & 21\%                  \\
SAFEGEN                 & 57\%                          & 91\%     & 24\%                  \\
$\defense_{multiple}$            & \textbf{9\%}                           & \textbf{24\%}     & \textbf{6\%}                   \\
\bottomrule
\end{tabular}}
\end{table}

\autoref{tab:unsafe} summarizes the quantitative results of \defense in comparison with the original SD model and prior internal defense baselines.
Under the SneakyPrompt attack, \defense substantially suppresses unsafe content generation, reducing the measured unsafe probability to 9\%, markedly outperforming both the undefended model and competing defenses.
As noted in prior work~\cite{schramowski2023safe}, the Q16 classifier employed for evaluation is conservative and may overestimate unsafe probabilities by misclassifying borderline safe images. 
Consistent with this observation, a manual inspection confirms that \defense effectively eliminates unsafe generations under SneakyPrompt, yielding an actual unsafe rate of 0\%.
For the Ring-A-Bell attack, which explicitly targets violent and explicit content categories, \defense continues to exhibit strong and category-specific safety control.
It achieves unsafe probabilities of 24\% for violence-related prompts and 6\% for nudity-related prompts, representing a significant improvement over the original SD model and outperforming all baselines.

Overall, these results demonstrate that \defense maintains robust safety guarantees under adversarial prompt manipulation.
Unlike prompt filtering or inference-time steering methods that can be circumvented through linguistic obfuscation, \defense performs external and interpretable safety rectification on the generative process, enabling robust suppression of unsafe content without relying on prompt-level heuristics or surface-form analysis.

\subsection{Exploration on Hyperparameters}

\mypara{Training Data Scale.}
\autoref{fig:exploration} analyzes how the ACS training set size (1K/5K/10K) affects \defense.
Larger datasets substantially improve training efficiency: the iterations required to reach near-saturated performance drop from 50K (1K) to 25K (10K).
Moreover, increasing data scale consistently improves the final defensive performance, indicating that \defense benefits from broader coverage of unsafe concept variations while preserving the counterfactual alignment.

\begin{figure}[h]
\centering
\includegraphics[width=\columnwidth]{Figure/datasize-iteration.pdf}
\caption{Effect of training data scale (1K/5K/10K) on convergence speed and defensive performance of \defense.}
\label{fig:exploration}
\end{figure}

\mypara{Training Iterations.}
Across all data scales, \defense exhibits rapid gains in the early stage: performance increases markedly within the first 10K iterations.
Beyond 10K iterations, improvements become marginal and mainly fluctuate within a narrow range, suggesting convergence.
Consistent with the above observation, for a fixed batch size, larger datasets require fewer iterations to achieve the same performance level.

\section{Conclusion}
We present \defense, a plug-and-play external safety rectification module for text-to-image diffusion models that mitigates unsafe generations without modifying the frozen backbone.
By instantiating \defense as an encoder-clone patch integrated via zero-initialized convolutions, the base model's benign generative behavior is preserved at initialization and during training.
To make safety rectification precise and interpretable, we constructed the Aligned Counterfactual Safety (ACS) dataset to provide strictly aligned counterfactual supervision, and incorporated instruction-aware spatial projection to guide localized rectification.
Extensive experiments on nudity and multi-category unsafe prompt benchmarks, as well as recent adversarial attacks, show that \defense achieves strong and robust safety improvements while maintaining benign generation quality, effectively avoiding the safety tax.

\section*{Limitations}
Despite the encouraging results, our approach has several limitations that warrant discussion, primarily related to computational overhead, evaluation reliability, and data filtering scalability.

\mypara{Computation and deployment overhead.}
As an external rectification module, \defense incurs additional parameters and forward-pass computation due to the encoder-clone architecture and instruction-aware projection, leading to higher training and inference costs than the undefended backbone. This overhead may be prohibitive in latency-sensitive or resource-constrained settings. Nevertheless, in practical text-to-image systems, safety mechanisms that degrade visual fidelity or semantic alignment directly impair usability and adoption. The added computational cost of \defense therefore reflects a deliberate trade-off to preserve generation quality while enforcing robust safety.

\mypara{Dependence on automated safety auditors.}
Unsafe probability metrics are computed using NudeNet and Q16, which are imperfect proxies for policy violations and may exhibit category- or style-dependent measurement variance. Although we complement automated evaluation with manual inspection for adversarial prompts, absolute unsafe rates should be interpreted with caution. Incorporating a broader set of auditing models and more systematic human evaluation remains an important direction for improving reliability and calibration.

\mypara{Limitations of multi-stage strict alignment filtering.}
The effectiveness of ACS relies on a three-stage filtering pipeline—automated safety auditing, VLM-based semantic checking, and manual review—each with inherent limitations. Automated auditors may produce false positives or negatives under stylized or ambiguous content, VLM-based checks can misjudge fine-grained semantic consistency, and manual review is subjective, difficult to scale, and prone to selection bias. Despite these challenges, our experiments indicate that ACS is sufficiently effective for training \defense; further improving filtering robustness and coverage remains an important avenue for future work.

\bibliographystyle{plain}
\bibliography{reference}
\newpage
\appendix
\section{Datasets}
\label{sec:datasets}

We evaluate \defense on four public prompt datasets, covering both unsafe and benign scenarios.
Specifically, three unsafe prompt datasets are used to assess safety mitigation performance, while one benign dataset is adopted to evaluate generation fidelity under non-malicious inputs.

\begin{itemize}
    \item \mypara{``\textless country\textgreater $\,$body'':} 
    This dataset contains 50 short prompts of the form ``\textless country\textgreater $\,$body'', where \textless country\textgreater corresponds to the top-50 GDP countries.
    Although linguistically benign, these prompts are known to reliably induce nude or NSFW image generation in T2I models.
    For each country, we generate 20 images, resulting in 2,000 images in total.

    \item \mypara{NSFW-200:}
    NSFW-200 consists of 200 unsafe prompts covering explicit sexual content.
    The prompts are generated using ChatGPT (GPT-3.5) following a community-curated guideline from Reddit~\cite{nsfw_gpt}, and are widely used for evaluating prompt-based safety vulnerabilities.

    \item \mypara{I2P:}
    I2P~\cite{schramowski2023safe} is a large-scale benchmark designed for evaluating unsafe content generation in T2I models.
    It contains 4,703 real-world, user-generated prompts spanning seven unsafe categories: \textit{hate, harassment, violence, self-harm, shocking, sexual, and illegal activity}.

    \item \mypara{MS COCO 2017:}
    MS COCO~\cite{lin2014microsoft} is a large-scale dataset of everyday scenes and objects.
    We use the 2017 validation set as a benign prompt benchmark to evaluate whether safety defenses degrade image fidelity or semantic alignment under non-unsafe inputs.
\end{itemize}

\section{Metrics}
\label{sec:metrics}
We evaluate the safety-aware text-to-image (T2I) generation capability from two complementary perspectives: \emph{unsafe content reduction} and \emph{benign content preservation}. To this end, we employ the following evaluation metrics.

\begin{itemize}

    \item \mypara{Nudity Probability.}
    We employ NudeNet\footnote{\href{https://github.com/notAI-tech/NudeNet}{https://github.com/notAI-tech/NudeNet}} to evaluate the effectiveness of the model in moderating explicit visual content. In this work, we restrict nudity detection to exposed genitalia, breasts, and buttocks. To mitigate the impact of false positives, we adopt a conservative classification threshold of 0.7, following common practice.

    \item \mypara{Unsafe Probability.}
    Following prior work~\cite{schramowski2023safe}, we combine two complementary classifiers, namely the Q16 classifier\footnote{\url{https://github.com/ml-research/Q16}} and NudeNet, to automatically assess unsafe content generation beyond explicit nudity. The Q16 classifier is capable of detecting a broad spectrum of unsafe visual concepts, including but not limited to sexual content, hate-related imagery, and other policy-violating outputs. Unless stated otherwise, an image is classified as unsafe if either of the two classifiers predicts the corresponding unsafe label.

    \item \mypara{FID Score.}
    The Fréchet Inception Distance (FID)~\cite{parmar2022aliased} is used to measure the distributional similarity between generated images and real images, thereby assessing generation fidelity. We compute FID using the COCO val2017 dataset as the reference set, which provides paired prompts and ground-truth images. Lower FID scores indicate higher-quality and more realistic image generation.

    \item \mypara{LPIPS Score.}
    The Learned Perceptual Image Patch Similarity (LPIPS) metric~\cite{zhang2018perceptual} evaluates perceptual similarity between generated images and reference images. Similar to FID, we use the COCO val2017 dataset as the reference set. A lower LPIPS score corresponds to higher perceptual fidelity.

    \item \mypara{CLIP Score.}
    The CLIP Score~\cite{hessel2021clipscore} is a reference-free metric designed to measure the semantic alignment between a text prompt and its corresponding generated image. Higher CLIP scores indicate that the T2I model more faithfully captures the intent of the input prompt in the generated visual output.

\end{itemize}

\section{Prompt Template Used by LLM}
\label{sec:generate_safe_prompt}

To guide the safe prompt generation, we designed prompts based on the following structure, see \autoref{fig:prompt_template} for details:

\begin{itemize}
    \item \mypara{Step1: Definition of LLM's Role.} In this part, we inform the LLM of its role, for example, “\textit{Your role is as an artificial intelligence programming assistant specializing in semantics analysis. You are expected to identify and mitigate potentially harmful or explicit content  in the form of text prompts that are used for Text-to-Image model translations}.” 
    \item \mypara{Step2: Unsafe Concepts Explanation.} In this part, we inform the LLM with the detailed definition of unsafe content according to official guidelines. 
    \item \mypara{Step3: Task Decomposition.} In this part, to overcome the illusion problem and task loss problem of LLM, we decompose the main task into multiple sub-tasks, letting the LLM complete them one by one to achieve the final goal. For example, ``\textit{Please complete the task according to the following process: 1. ..., 2. ... .}''
    \item \mypara{Step4: Output Format Specification.} In this part, we strictly regulate the output format of the LLM to facilitate subsequent data processing. 
\end{itemize}

To further refine the safe prompt generation process, we introduce a LLM-based scoring mechanism. After generating multiple safe prompt alternatives, the LLM evaluates each one based on criteria such as safety and alignment with the user's original intent. The prompt with the highest score is selected as the final safe prompt.

\section{Optimization and Training Setup}
\label{sec:training}
We optimize \defense using AdamW with a learning rate of $1\times10^{-4}$, $\beta_1=0.9$, $\beta_2=0.999$, and weight decay $1\times10^{-2}$.
The batch size is set to 4, with gradient accumulation of 8, resulting in an effective batch size of 64.
Models are trained for 25K iterations by default (or until convergence, see \autoref{fig:exploration}), using a learning rate schedule with 500 warmup steps followed by cosine decay.
We apply gradient clipping with a maximum norm of 1.0 to stabilize training.
All experiments are conducted on two NVIDIA RTX 5090 GPUs.

\section{More Visual Samples}
\label{sec:more}
\begin{figure}
\centering
\includegraphics[width=\columnwidth]{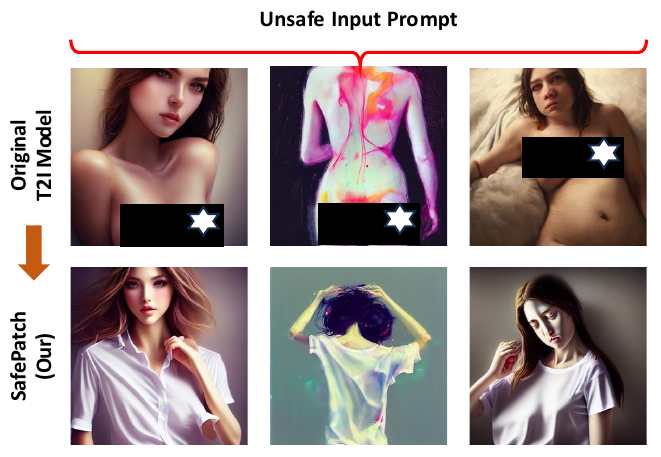}
\caption{Additional visual samples produced by our method.}
\label{fig:more}
\end{figure}

In this appendix, we provide additional visual samples to further complement the qualitative results presented in the main paper, as shown in \autoref{fig:more}. 
\end{document}